\pdfoutput=1
\documentclass[journal]{IEEEtran}

\usepackage{xcolor,soul,framed} 

\colorlet{shadecolor}{yellow}
\usepackage[pdftex]{graphicx}
\graphicspath{{../pdf/}{../jpeg/}}
\DeclareGraphicsExtensions{.pdf,.jpeg,.png}

\usepackage[cmex10]{amsmath}
\usepackage{array}
\usepackage{mdwmath}
\usepackage{mdwtab}
\usepackage{eqparbox}
\usepackage{url}

\usepackage{amsmath} 
\usepackage{amssymb}  
\usepackage{cite} 
\usepackage{color}

\usepackage{algorithm}
\usepackage{algorithmicx}
\usepackage{algpseudocode}
\usepackage{amsmath} 
\floatname{algorithm}{Algorithm} 
\let\labelindent\relax

\usepackage{multirow}
\usepackage{float} 

\usepackage{siunitx}

\usepackage{subfigure} 
\usepackage{graphicx}
\usepackage{enumitem}
\usepackage{todonotes}
\usepackage{makecell}
\usepackage{booktabs}
\usepackage{tablefootnote}
\begin{document}
    \title{Partial-to-Full Registration based on Gradient-SDF for Computer-Assisted Orthopedic Surgery}
    
\author{Tiancheng Li$^{1}$, Peter Walker$^{2}$, Danial Hammoud$^{2}$,  Liang Zhao$^{3}$, and Shoudong Huang$^{1}$ 
  
  \thanks{This work was supported by the National Health and Medical Research Council (NHMRC) Ideas Grant, Australia (No. 2029811).}
  \thanks{$^{1}$Tiancheng Li and Shoudong Huang  are with the Robotics Institute, Faculty of Engineering and Information Technology, University of Technology Sydney (UTS), Australia (e-mail: tiancheng.li-1@student.uts.edu.au; shoudong.huang@uts.edu.au).}
  
  \thanks{$^{2}$Peter Walker and Daniel Hammoud are with the Concord Repatriation General Hospital, New South Wales, Australia.}

  \thanks{$^{3}$Liang Zhao is with the School of Informatics, The University of Edinburgh, United Kingdom (e-mail: liang.zhao@ed.ac.uk).}

  \thanks{$^{*}$https://github.com/utsTianchengLi/Bone-registration-for-CAOS}
  }

\maketitle

\begin{abstract}
In computer-assisted orthopedic surgery (CAOS), accurate pre-operative to intra-operative bone registration is an essential and critical requirement for providing navigational guidance. This registration process is challenging since the intra-operative 3D points are sparse, only partially overlapped with the pre-operative model, and disturbed by noise and outliers. The commonly used method in current state-of-the-art orthopedic robotic system is bony landmarks based registration, but it is very time-consuming for the surgeons. To address these issues, we propose a novel partial-to-full registration framework based on gradient-SDF for CAOS. The simulation experiments using bone models from publicly available datasets and the phantom experiments performed under both optical tracking and electromagnetic tracking systems demonstrate that the proposed method can provide more accurate results than standard benchmarks and be robust to 90\% outliers. Importantly, our method achieves convergence in less than 1 second in real scenarios and mean target registration error values as low as 2.198 $mm$ for the entire bone model. Finally, it only requires random acquisition of points for registration by moving a surgical probe over the bone surface without correspondence with any specific bony landmarks, thus showing significant potential clinical value. The MATLAB code of the framework is available$^{*}$.


\end{abstract}


%


\section{Introduction}

\IEEEPARstart{O}{steoarthritis} constitutes a major musculoskeletal burden for aged people worldwide. The demand for computer-assisted orthopedic surgery (CAOS) for treating osteoarthritis has grown substantially, especially total hip replacements (THR) and total knee arthroplasty (TKA). The bone registration is a fundamental requirement for providing navigational guidance in CAOS \cite{sugano2003computer}. Its main task is to bring the pre-operative computerised tomography (CT) or magnetic resonance imaging (MRI) space and the intra-operative patient space together \cite{min20233d} by solving the transformation matrix between two bone point sets \cite{ma2003robust}, which is necessary for accurate surgical planning and placement of implants. The pre-operative model (full bone point sets) which is typically global and complete, and can be obtained by segmenting the bone from the CT or MRI volume \cite{chen2023generalized}. The intra-operative partial point set is usually obtained from medical devices such as optical tracking system, endoscopic camera, electromagnetic tracking, CT or X-ray during the procedure \cite{ren2010investigation, wang2018surgical, li2023closed, zhang2022slam, gopalakrishnan2024intraoperative}. 

\begin{figure}[h]
	\centering
	\includegraphics[width=1\linewidth]{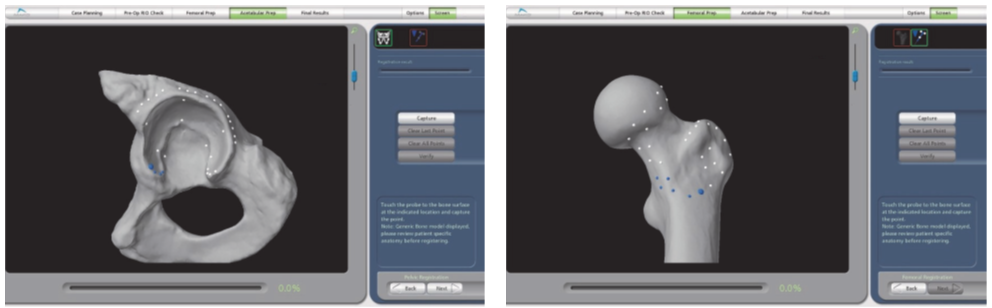}
	\caption{Registration pattern for pelvis and femur in MAKO \cite{lonner2019robotics}. The dots show the bony landmarks to be accurately verified and acquired by the probe for registration.}
	\label{fig:mako}
\end{figure}

The most natural and commonly used registration method is to use a surgical tracking probe to verify and acquire multiple bony landmarks intra-operatively corresponding to the pre-operative model \cite{lonner2019robotics}. However, it is difficult to accurately recognize and touch the exact landmarks during surgery due to blood and surface tissue interference \cite{min2020robust}. Errors in landmark selection can also severely affect the accuracy and robustness of registration. Therefore, the current robotic systems for orthopedic surgery require the surgeon to select as many bony landmarks as possible for registration to increase error tolerance and improve the registration accuracy. For example, the current state-of-the-art orthopedic robotic system, MAKO Robotic Arm Interactive Orthopedic System (Stryker), needs more than 30 landmarks to be accurately touched by the probe as shown in Fig. \ref{fig:mako}. This approach will lead to an increase in the average surgery time and potential registration error due to the possible invisibility of the landmarks during surgery \cite{lonner2019robotics}. 

A better alternative is to randomly acquire points by moving the surgical tracking probe over the bone surface intra-operatively and align with the pre-operative 3D model. However, there are multiple challenges: (i) As only part of the bone can be exposed intra-operatively, the acquired points could only partially overlap with the pre-operative model. The partial anatomy can lead to limited spatial and structural information for the registration; (ii) Intra-operatively acquired points are usually disturbed by noise and outliers. The low-quality point sets can affect the accuracy of the registration. (iii) The intra-operative points are sparse, resulting in minimal structural features and increasing the registration difficulty.  

In this paper, we propose a robust and accurate partial-to-full registration approach based on gradient signed distance field (gradient-SDF) \cite{sommer2022gradient} to address the above mentioned challenges. The simulation experiments on multi-category bone structures from publicly available datasets, and real phantom experiments under both optical tracking and electromagnetic tracking systems show the competitiveness of the proposed framework: (i) Gradient-SDF can provide more accurate gradients so that it ensures accurate registration and fast convergence for sparse point clouds without the need for correspondences. (ii) Using M-estimator with Cauchy’s residual function can effectively remove outliers and improve robustness to noise.  (iii) Our approach only require random acquisition of points by moving the probe over the bone surface, which can lead to saving of surgical time and cost. (iv) It can be easily integrated clinically, without interrupting the workflow of COAS. 

\section{Related Work}

\textbf{Traditional registration methods.} The Iterative Closest Point (ICP) algorithm \cite{besl1992method} is considered a milestone in point cloud registration, which is an iterative algorithm estimating the pose and correspondence simultaneously. However, ICP is prone to converge to local minima and only performs well given an accurate initial guess \cite{yang2020teaser}. Multiple variants of ICP have been proposed, including those using normal information \cite{ma2003robust,serafin2015nicp}, incorporating uncertainty in measurements \cite{segal2009generalized} or finding correspondence \cite{billings2015iterative}. To further improve the robustness and accuracy of the point cloud registration, methods based on probabilistic interpretations have been proposed. For example, Gaussian Mixture Model (GMM) \cite{jian2010robust} represents two point sets in search of the optimal transformation. Coherent Point Drift (CPD) \cite{myronenko2010point} is also one classic method in this category, where one point set is considered as the centers of the Gaussian mixtures and the other point set is generated by the mixtures. JRMPC \cite{evangelidis2014generative} generalises CPD to the case where multiple point sets are registered simultaneously. Motivated by getting the globally optimal solution, Go-ICP \cite{yang2015go} adopts a branch and bound technique to search the entire 3D motion space. TEASER \cite{yang2020teaser} reformulates the registration problem using a Truncated Least Squares (TLS) cost and solves it by a general graph-theoretic framework.

\textbf{Learning-based registration methods.} The breakthroughs in deep learning for 3D point clouds (e.g., PointNet \cite{qi2017pointnet} and DGCNN \cite{wang2019dynamic}) present new prospects for learning point cloud registration directly from data. PointNetLK \cite{aoki2019pointnetlk} uses PointNet to learn feature representation, and then align source and target point set features, thereby enabling them to be aligned. However, PointNetLK is mainly proposed for point set pairs with a high ratio of overlapping. To address the challenge of partial overlap, PRNet \cite{wang2019prnet} extends DCP \cite{wang2019deep} to aligning partially overlapped point clouds by employing a Siamese DGCNN \cite{wang2019dynamic} for feature embedding. Nevertheless, these methods that perform well are overwhelmingly based on correspondence and are difficult to ensure generalization when transferred to novel registration tasks \cite{chen2023generalized}.

\textbf{Registration in CAOS.} Depending on the data type, point set registration problems in CAOS can be classified into 2D-to-3D registration and 3D-to-3D registration problems \cite{taylor2020medical}. Focusing on our 3D-to-3D rigid point set registration case, the most commonly used method is the landmark-based registration \cite{lonner2019robotics}. It is also the current method used by the MAKO robot. Lots of  anatomical landmarks of interest need to be precisely localised in both the pre-operative and intra-operative spaces, which is time consuming and error-prone \cite{min20233d}. Motivated by the use of a small number of sparse measurements probing-based surgical registration, Srivatsan \emph{et al.} \cite{arun2019registration} propose two variants for sparse point and normal registration (SPNR), namely deterministic (dSPNR) and probabilistic (pSPNR). However, as the name suggests, the methods require surface normal measurements for the intra-operative acquired points. Although the authors suggest that real scenarios can end up with a robotic arm to help obtain the normal vector information, this would also add cost and time. Min \emph{et al.} \cite{min2020robust} utilized the tangent vectors extracted from the sparse intra-operative data points and the normal vectors extracted from the pre-operative model points for registration, which also increases the surgeon's intra-operative workload. 

This paper aims to address the above challenges and issues. The main contributions can be summarized as:
\begin{itemize}
	\item We propose a novel and robust framework based on gradient-SDF for a challenging scenario: registration from a sparse and partial point set to another full model. 	
	\item The proposed method does not require correspondence, but only random acquisition of points by moving a probe over the bone surface, thus simplifying the surgeon's work during intra-operative registration and showing significant potential clinical value.       
    \item Experiments on multi-category sawbones models were carried out in both optical and electromagnetic tracking systems to verify that our framework can obtain accurate registration results in real scenarios with very fast convergence rates.
\end{itemize}

\section{METHODOLOGY}
\begin{figure}[t]
	\centering
	\includegraphics[width=1\linewidth]{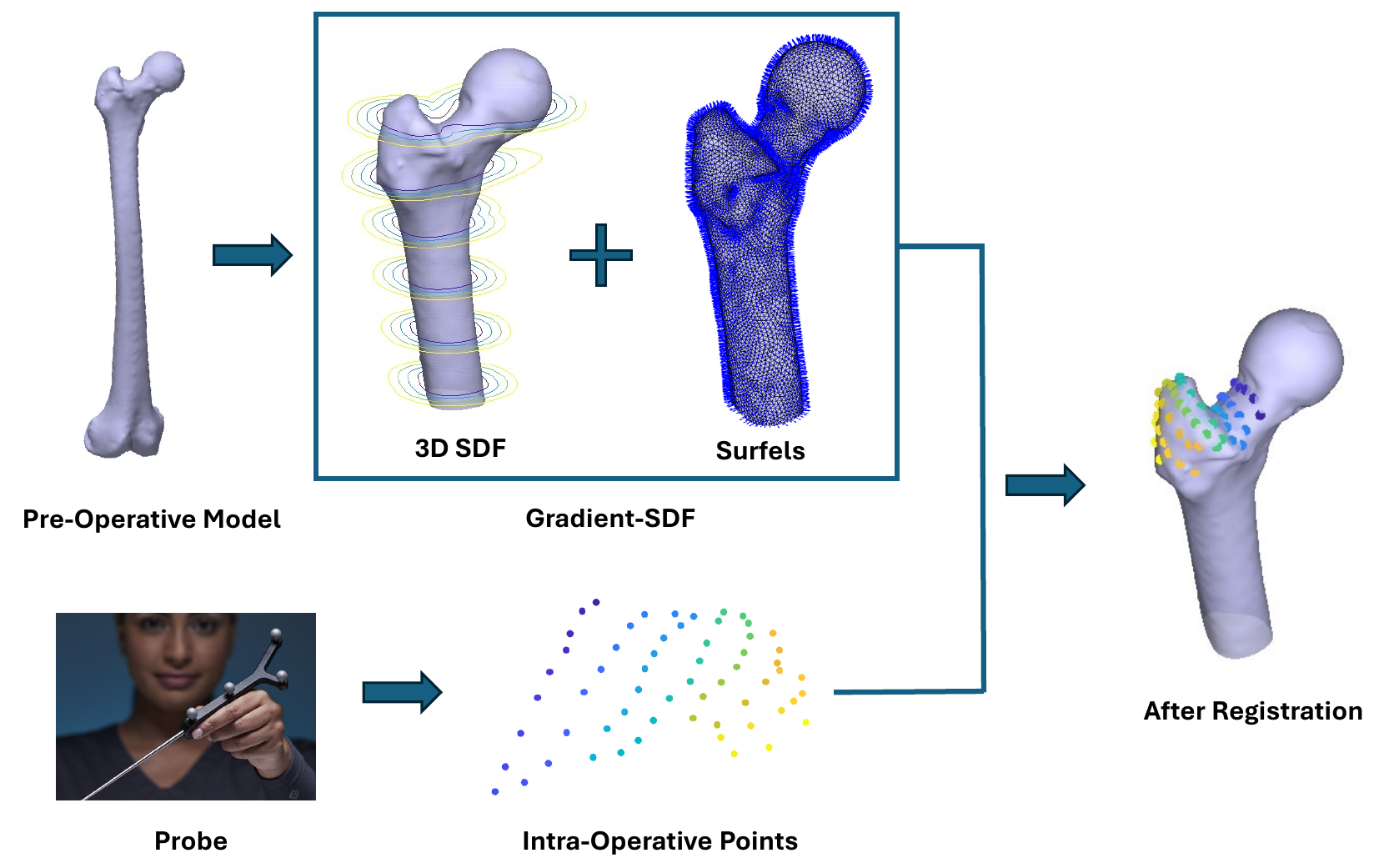}
	\caption{Our framework: the pre-operative model is represented as gradient-SDF which is a hybrid representation between standard SDF stored in a voxel grid and explicit geometry representation using surfels (surface normal). The intra-operative points are collected by moving the probe over the bone surface.}
	\label{fig:framework}
\end{figure}
\subsection{Optimisation Problem Formulation}
For the standard CAOS, the pre-operative model is obtained from CT scan or MRI. The model is the full 3D point set, and the first step in our algorithm is to represent it as gradient-SDF which is a voxel grid where each voxel contains its Euclidean distance to the nearest object surface (Fig. \ref{fig:framework}). Different from the standard SDF, gradient-SDF also pre-calculates and stores the gradients for each voxel using the surface normal property of a surfel \cite{sommer2022gradient}. 

The sparse point cloud $p_k (\ k\in\{1, \cdots, K\}$) is randomly acquired by moving the tracking probe on the bone surface intra-operatively (Fig. \ref{fig:framework}). The goal is to transform this point cloud to the surface of the model, that is, the zero set level of the gradient-SDF such that $SDF(p)=0$, where $p \in \mathbb{R}^3$ is a point in the space. Then, the problem is estimating a rigid motion with rotation $\mathbf{R} \in SO(3)$ and the translation $\mathbf{t} \in \mathbb{R}^3$, which minimises the following function: 
\begin{equation}\label{Eq:Rt_LS}
    F(X) = \sum_{k=1}^K \|{ f_k(X)\|^2} = \sum_{k=1}^K \|{SDF(\mathbf{R} \cdot p_k + \mathbf{t})\|^2.}
\end{equation}
\noindent where $X=(\mathbf{R}, \mathbf{t})$ is the state, and $ f_k(X)$ is the closest distance from  $p_k$ to the object surface. Applying Gauss-Newton method, for each iteration with initial state $X^{(n)}$, the optimal update $\Delta X^*$ can be obtianed by
\begin{equation}\label{Eq:loss}
     \Delta X^*=\mathop{\arg \min}\limits_{\Delta X} \sum_{k=1}^K \|{SDF(P_k(X^{(n)} \oplus \Delta X))\|^2},
\end{equation}
\noindent where 
\begin{equation}
  \begin{split}
    P_k(X) &=\mathbf{R} \cdot p_k + \mathbf{t}, \\
    X \oplus \Delta X &= (\text{exp}({\xi}^{\wedge}) \cdot {\mathbf{R}},\mathbf{t}+\phi),
  \end{split}
\end{equation}
and $\Delta X=[\xi, \phi]\in \mathbb{R}^6$ is the update increment. ${(\cdot)^{\wedge}}$ is the skew-symmetric operator, and $\text{exp}(\cdot)$ is the matrix exponential map.
Then, the state $X^{(n+1)}$ is updated by
\begin{equation}
    X^{(n+1)}=X^{(n)}\oplus \Delta X^*.
\end{equation}
When the algorithm converges, we can obtain the solution of problem (\ref{Eq:Rt_LS}).




In particular, the Jacobians in the above Gassu-Newton iterations are calculated as 
\begin{equation}\label{Eq:lie_Ji}
     J_k(X) = \frac{\partial f_k(X)}{\partial X} = {\nabla}_{SDF}(P_k(X))  \cdot \frac{\partial P_k(X)}{\partial X}, 
\end{equation}
\noindent where ${\nabla}_{SDF}(.)$ represent the gradient, which is equal to the inwards-pointing surface normal at the closest surface point, and the negative of the outwards-pointing surface normal \cite{sommer2022gradient}. And 
\begin{equation}\label{Eq:lie_Jp}
     \frac{\partial P_k(X)}{\partial X}= [-(\mathbf{R} \cdot p_k)^{\wedge}, \mathbf{I}_{3\times3} ]. 
\end{equation}

Solving the formulated optimisation problem using the above method is computationally efficient since the distance space $SDF(.)$ and the gradient ${\nabla}_{SDF}(.)$ can be pre-calculated on the pre-operative data. This gradient naturally contains the geometric properties that are not present in the gradients calculated by standard SDF, so the convergence is better and faster.








\subsection{Robust M-estimator against Noise and Outliers}
The robustness of the method is crucial since the intra-operative acquired partial points may also be susceptible to noise and outliers due to interference between medical devices. The robust M-estimators (maximum likelihood type estimators) are originally used to reduce the effect of outliers and to find the maximum likelihood estimate and now have been used in many simultaneous localisation and mapping (SLAM) applications \cite{hu2013towards}. In this work, we propose to loop over M-estimator with a re-descending influence function to remove the outliers and against the noise. The idea behind it is to design and solve a sequence of intermediate optimization problems $\mathcal{P}_1,...,\mathcal{P}_N$ so that (i) the solution of each problem $\mathcal{P}_i$ can be used as the initial guess to obtain the maximum likelihood estimate in the next problem $\mathcal{P}_{i+1}$, and (ii) the solution of the final problem $\mathcal{P}_N$ is within the basin of attraction of the global minimum of the original least squares problem (\ref{Eq:Rt_LS}).

Similar to the idea of Graduated Non-Convexity \cite{blake1987visual}, $\{\mathcal{P}_i\}_{i=1}^N$ can be defined as assigning additional weights $w_k^{(i)}$ to all the acquired points $p_k$ such that measurements with large residuals in $\mathcal{P}_i$ will update smaller weights for $\mathcal{P}_{i+1}$. So the $i$th intermediate optimization problem $\mathcal{P}_i$ can be formulated as extending the (\ref{Eq:Rt_LS}) to
\begin{equation}\label{Eq:weight_loss}
    X^{\star}_{(i)}=\mathop{\arg \min}\limits_{X} \sum_{k=1}^K w_k^{(i)} \|{ f_k(X)\|^2}. 
\end{equation}
\noindent This is similar to the iterated re-weighted least squares (IRLS) problem formulated in M-estimators to reduce the effect of outliers \cite{zhang1997parameter}. Large residuals in this problem correspond to the potential outliers or noisy acquired points. As the iterations proceed, the weights assigned to the parts with large residuals become smaller and smaller until they are negligible.

\subsection{Selection of the M-estimator}
Instead of the squared residuals, the M-estimators apply another function of the residuals, yielding 
\begin{equation}\label{Eq:residual}
    \min \sum_{k=1}^K \rho (e_k), 
\end{equation}
\noindent where $e_k \triangleq \| f_k(X)\|_2$ denotes the ${\ell _2}$-norm of $f_k(X)$ and $\rho (.)$ is a symmetric positive-definite function with a unique minimum at zero with increments less than squared residuals. In this paper, we use Cauchy's function, since it is more stable against different environment types \cite{babin2019analysis}. For the Cauchy's function we have \cite{zhang1997parameter}:
\begin{equation}\label{Eq:cauthy}
    \rho (e_k) \triangleq \frac{c^2}{2} \log (1 + {(\frac{e_k}{c})}^2).
\end{equation}

Equation (\ref{Eq:residual}) can be implemented as an IRLS problem by computing the gradient w.r.t. $X$ and setting it to zero:
\begin{equation}\label{Eq:re-formulate}
    \sum_{k=1}^K \psi(e_k) \frac{\partial e_k}{\partial X} =0,
\end{equation}
\noindent where $\psi(x) \triangleq d\rho (x) / dx$ is the influence function and the weight function can be defined as $w(x) \triangleq \psi(x) / x$, then (\ref{Eq:re-formulate}) can be re-written as:
\begin{equation}\label{Eq:re-w}
    \sum_{k=1}^K w(e_k)e_k \frac{\partial e_k}{\partial X} =0.
\end{equation}

This is the exact system of equations that results from solving the following IRLS problem
\begin{equation}\label{Eq:IRLS}
    \min \sum_{k=1}^K w(e_k^{(i-1)})e_k^2,
\end{equation}

\noindent where the superscript $i$ indicates the iteration number, $e_k^{(i-1)}$ is the residual calculated using the latest estimate, and the weight $w(e_k^{(i-1)})$ need to be re-computed after each iteration for the next estimation. It is noted that (\ref{Eq:IRLS}) is consistent with the underlying idea of (\ref{Eq:weight_loss}) and the performance depends on the selection of the M-estimator influence function $\psi (.)$. In our cases, the rate of descent for Cauchy M-estimator (\ref{Eq:cauthy}) is:
\begin{equation} \label{Eq:rate}
  \begin{split}
  \psi(e_k) \triangleq \frac{e_k}{1+e_k^2},  \qquad
  w (e_k) \triangleq \frac{1}{1+e_k^2}.
  \end{split}
\end{equation}

After the convergence of Gauss-Newton iteration, the updated weights truly reflect the contribution of each acquired point. The points with small weights are outliers and needed to be eliminated. Then repeat IRLS optimisation process with the inliers. This robust registration approach is summarized in Algorithm \ref{alg:cap}.   

\begin{algorithm}[t]
\caption{Robust registration framework based on gradient-SDF and M-estimator}\label{alg:cap}
\begin{algorithmic}[1]
\Require Full point set $\mathbf{P}$; sparse point set: $\{p_k\}_{k=1}^K$. 
\Ensure Optimal transformation $X^{\star}$. 

\State Represent $\mathbf{P}$ as gradient-SDF. 
\State Initial transformation $X^{(0)}$.
\State $X^{\star}_{(0)} \leftarrow X^{(0)}$

\Loop
\Repeat
\For {each $p_k$}
   \State Calculate $e_k^{(i-1)}$ using $X^{\star}_{(i-1)}$ by (\ref{Eq:Rt_LS}) and (\ref{Eq:residual})
   \State $w_k^{(i)} = w(e_k^{(i-1)})$  // $Cauchy$ $M$-$estimator$
\EndFor
    \State Calculate $X^{\star}_{(i)}$ by solving the optimization in (\ref{Eq:weight_loss}) with $X^{\star}_{(i-1)}$ as initial value
    \State $i \leftarrow i+1$
\Until { $\|w^{(i)}-w^{(i-1)}\| \leq \varepsilon$ } // $Converge$

\If {exist $w^{(i)}_k < \delta $}
    \State Discard the corresponding $p_k$ // $Outliers$
    \State $\{p_k\}_{k=1}^K \leftarrow \{p_k\}_{k=1}^{K^{\star}}$ and continue the loop
\Else
    \State $X^{\star} = X^{\star}_{(i)}$
\EndIf

\EndLoop
\end{algorithmic}
\end{algorithm}

\section{Experiments and Results}

\subsection{Experimental datasets}
\textbf{Bone structure datasets.} In this study, we selected several bone surface data from publicly available datasets that are all created from CT scans.\cite{fischer2020robust, fischer2023database, keast2023geometric}. RWTHmediTEC \cite{fischer2020robust} only obtains femur and hip structures. VSDFull \cite{fischer2023database} includes the entire lower body which is a complementary version of RWTHmediTEC, and SSM-Tibia \cite{keast2023geometric} only has the right tibia. The detailed composition of the dataset is reported in Table \ref{tab: dataset}. For example, RWTHmediTEC contains 19 right and 18 left femurs, and 20 each of the left and right hip structures.

\begin{table}[t]
\centering
\caption{Summary of the bone structure datasets. L and R in table stand for left and right, respectively.}\label{tab: dataset}
\begin{tabular}{lllllll}
\toprule
\multicolumn{1}{l}{\multirow{2}{*}{Dataset}} & \multicolumn{2}{c}{Femur} & \multicolumn{2}{c}{Hip} & \multicolumn{2}{c}{Tibia} \\ \cline{2-7} 
\multicolumn{1}{c}{}                         & R           & L           & R            & L           & R           & L           \\ 
\midrule
RWTHmediTEC \cite{fischer2020robust}                                            & 19          & 18          & 20           & 20          &     -        &      -       \\ 
VSDFull \cite{fischer2023database}                                            & 30          & 30          & 30           & 30          & 30          & 30          \\ 
SSM-Tibia \cite{keast2023geometric}                                            &      -       &       -      &         -     &   -          & 30          &     -        \\ 
\bottomrule
\end{tabular}
\end{table}

\begin{figure}[h]
	\centering
	\includegraphics[width=1\linewidth]{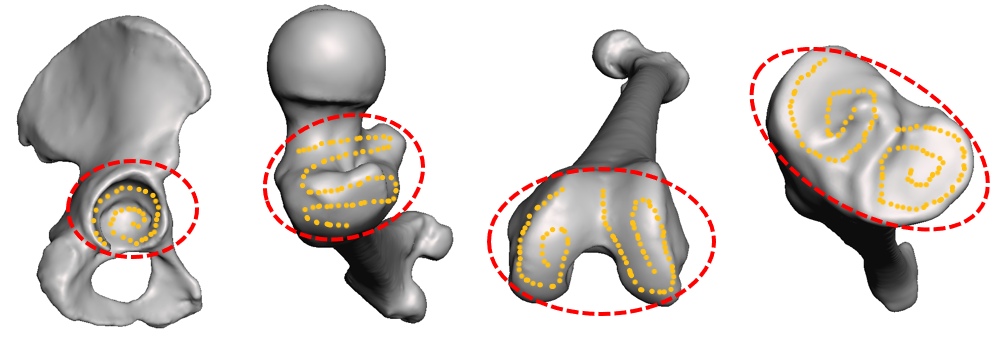}
	\caption{The parts circled in red dotted lines are bone structures that may be exposed during orthopaedic surgeries. In order from left to right: pelvic acetabulum; proximal femur; femoral condyle; proximal tibia. The yellow points are the simulated surgical probe picking points, which are used as partial data to be registered with the full point cloud model.} 
	\label{fig:partial}
\end{figure}

\textbf{Partial point sets generation}. The partial point sets were generated from the models, and shaped as curved to simulate the acquired points by moving the surgical probe (refer to the yellow dots in Fig. \ref{fig:partial}). The points were distributed only on the proximal femur and acetabulum exposed in THR, and on the proximal tibia, medical and lateral femoral condyle exposed in TKA (Fig. \ref{fig:partial}). We applied a rigid random transformation to the partial point sets, involving a rotation $R$ on each axis within -45 to 45 degrees in each of three axises, and a translation $t$ within -1000 $mm$ to $1000mm$ for each element to serve as the ground truth. 

\begin{table*}[thbp]
\centering
\caption{Registration Results on Various Bone Datasets Using Different Methods}\label{tab: compare}
\begin{tabular}{llcccccccccccc}
\toprule
\multicolumn{1}{c}{\multirow{2}{*}{Noise Type}} & \multicolumn{1}{c}{\multirow{2}{*}{Methods}} & \multicolumn{3}{c}{Hip}                                                                                                                                            & \multicolumn{3}{c}{Proximal Femur}                                                                                                                                  & \multicolumn{3}{c}{Femoral Condyle}                                                                                                                                & \multicolumn{3}{c}{Proximal Tibia}                                                                                                                                 \\ \cline{3-14} 
\multicolumn{1}{c}{}                            & \multicolumn{1}{c}{}                         & \begin{tabular}[c]{@{}c@{}}MAE\\ (R/deg)\end{tabular} & \begin{tabular}[c]{@{}c@{}}MAE \\ (t/mm)\end{tabular} & \begin{tabular}[c]{@{}c@{}}CD \\ (mm)\end{tabular} & \begin{tabular}[c]{@{}c@{}}MAE \\ (R/deg)\end{tabular} & \begin{tabular}[c]{@{}c@{}}MAE \\ (t/mm)\end{tabular} & \begin{tabular}[c]{@{}c@{}}CD \\ (mm)\end{tabular} & \begin{tabular}[c]{@{}c@{}}MAE\\ (R/deg)\end{tabular} & \begin{tabular}[c]{@{}c@{}}MAE \\ (t/mm)\end{tabular} & \begin{tabular}[c]{@{}c@{}}CD \\ (mm)\end{tabular} & \begin{tabular}[c]{@{}c@{}}MAE\\ (R/deg)\end{tabular} & \begin{tabular}[c]{@{}c@{}}MAE \\ (t/mm)\end{tabular} & \begin{tabular}[c]{@{}c@{}}CD \\ (mm)\end{tabular} \\ 
\midrule
\multirow{6}{*}{Isotropic} & ICP    & 42.102  & 4.672  & 5.208  & 21.865  & 93.490  & 4.344  & 47.69  & 99.024  & 9.077  & 64.320  & 80.729  & 4.986   \\
 & CPD    & 21.327  & 9.855  & 6.384 &30.553   & 95.697 & 4.414  & 59.811   & 108.143  & 8.070  & 60.485  & 72.098  & 4.855  \\
& JRMPC   &18.932	&3.321	&3.231	&22.735	&80.523	 &4.163	 &21.638	&91.427	&8.142	&41.648	 &73.831	&4.073\\
& Go-ICP   &2.131 &0.269 &0.765 &2.837 &2.024 &1.008 &1.242 &0.948 &1.071 &\textbf{1.125} &1.564 &0.804   \\
& Proposed   & \textbf{0.204}  & \textbf{0.202} & \textbf{0.706}     & \textbf{0.518}  & \textbf{0.667}   & \textbf{0.846}  & \textbf{0.605} & \textbf{0.473}  & \textbf{0.904}  & 1.127   & \textbf{0.763}  & \textbf{0.736} \\ 
\midrule
\multirow{6}{*}{Anisotropic} & ICP    &43.743	&17.868	&5.289	&22.358	&92.215	&4.398	&49.496	&98.201	&9.013	&65.151	&79.499	&4.789   \\
& CPD &21.853	&29.376	&6.349	&31.701	&96.082	&4.567	&59.695	&109.572	&8.186	&59.957	&72.367	&4.976 \\
& JRMPC  &19.343 &23.275	&3.458	&23.748	&80.145	 &4.124	 &23.953	&96.341	&9.742	&40.273	&73.164	&4.267 \\
& Go-ICP   &3.718	&0.403	&0.807	&3.154	&3.037	&1.005	&1.780	&0.431	&0.979	&6.153	&2.857	&0.846  \\
& Proposed  & \textbf{0.752}  & \textbf{0.177} & \textbf{0.691}      & \textbf{0.731}  & \textbf{0.557} & \textbf{0.889}   & \textbf{0.63}   & \textbf{0.408}  & \textbf{0.952}  & \textbf{1.304} & \textbf{0.945} & \textbf{0.811}    \\ 
\bottomrule
\end{tabular}
\end{table*}

\subsection{Comparison with benchmarks} \label{sec.compare}
In this study, we compared our proposed method with several benchmarks including ICP \cite{besl1992method}, CPD \cite{myronenko2010point}, JRMPC \cite{evangelidis2017joint}, and Go-ICP \cite{yang2015go}. Optical tracking and electromagnetic tracking are the most commonly used intra-operative medical devices to acquire partial point set. The accuracy (root mean square error) of optical tracking and electromagnetic tracking are 0.15 $mm$ and 0.7 $mm$, respectively \cite{sorriento2019optical}. Therefore, the zero-mean Gaussian noises were injected into the acquired partial point sets (refer to Fig. \ref{fig:partial}) to simulate the noise interference during acquisition. In terms of \emph{isotropic} positional noise, the standard deviation is $[0.5, 0.5, 0.5]$ $mm$, while $[0.3, 0.5, 0.7]$ $mm$ under \emph{anisotropic} positional noise. The quantitative metrics we used to evaluate the effectiveness of the proposed registration method were mean absolute errors (MAE) over Euler angles of $R$ and translation vector $t$, as well as Chamfer Distance (CD) which is the average distance between each point in one set to its nearest neighbor found in another \cite{lin2023hyperbolic}.

Table \ref{tab: compare} shows the comparison results by using different methods for various bone structures. The best results are bolded. Due to the excessive distribution disparity between the full and partial point sets, ICP, CPD and JRMPC performed poorly. Especially for slender bones such as the femur and tibia, these methods can easily fall into a local minima where the intra-operative sparse points converge to the middle of the femoral or tibial shaft, even given a good initial guess. Go-ICP performed much better when the mean squared error (MSE) convergence threshold is set to a very small value ($1e\mbox{-}5$). Our approach can always achieve the lowest MAE over Euler angles and translation vector in most test cases. We also try to get results from TEASER \cite{yang2020teaser}. Unfortunately, TEASER is not able to handle two point clouds that are disparate in number and have a tiny fraction of overlap.

\subsection{Resilience to noise and outliers}

In orthopedic surgery, the intra-operatively acquired points are usually disturbed by noise and the low-quality point sets may affect the accuracy of the registration. Therefore, it is necessary to verify the robustness of the proposed method in the presence of noise. We injected different levels of Gaussian noise into the acquired point set. The \emph{isotropic} positional noise levels increased gradually from 0.5 $mm$ to 1.2 $mm$. The standard deviation of \emph{anisotropic} positional noise increased gradually from $[0.3, 0.5, 0.7]$ $mm$ to $[1.0, 1.2, 1.4]$ $mm$. The registration results in Table \ref{tab: noise} are integrated for datasets of various bone types and demonstrate the resilience of our method in the face of noise disruptions. The proposed method performs well and converges fast under reasonable noise.

\begin{table}[thbp]
\centering
\caption{Registration Results on different noise level}\label{tab: noise}
\begin{tabular}{lccccc}
\toprule
\multicolumn{1}{c}{Noise Type} & \begin{tabular}[c]{@{}c@{}}Noise Levels \\ (mm)\end{tabular} & \begin{tabular}[c]{@{}c@{}}MAE \\ (R/deg)\end{tabular} & \begin{tabular}[c]{@{}c@{}}MAE \\ (t/mm)\end{tabular} & \begin{tabular}[c]{@{}c@{}}CD \\ (mm)\end{tabular} &\begin{tabular}[c]{@{}c@{}}Time \\ (s)\end{tabular} \\ 
\midrule
\multirow{5}{*}{Isotropic}  &0.5	&0.614	&0.526	&0.798	&0.077 \\
&0.7	&0.670	&0.782	&0.836	&0.151\\
&0.9	&1.021	&0.976	&1.057	&0.441\\
&1.2	&1.168	&1.029	&1.201	&0.653\\
\midrule
\multirow{5}{*}{Anisotropic} & [0.3,0.5,0.7]	&0.854	&0.522	&0.836	&0.313\\
& [0.5,0.7,0.9]	&0.674	&0.797	&0.895	&0.774 \\
& [0.7,0.9,1.1]	&1.077	&0.676	&1.054	&0.59
\\
& [1.0,1.2,1.4]	&1.575	&0.944	&1.169	&0.674
\\ 
\bottomrule
\end{tabular}
\end{table}

The intra-operative acquired partial points may also be susceptible to outliers due to interference between medical devices. To address this issue, we validated our proposed method by injecting various ratios of outliers $\{10\%, 30\%, 50\%, 70\%,90\% \}$ into the partial point sets. The injected outliers were randomly distributed around the bone model and the ratios are calculated as $N_{out}/(N_{in}+N_{out})$. The positional noise injected is the same as Section \ref{sec.compare}. As shown in Table \ref{tab: outlier}, the  proposed method is robust to increasing injected outliers up to $90\%$. As it is highlighted in bold, the largest rotational and translation error values are around 1.1 degree and 0.7 $mm$, which is still acceptable for a typical orthopedic surgery. The proposed method also performed well in terms of convergence speed, taking no more than $0.5$ seconds except for the extreme case of $90\%$ outliers. 

\begin{table}[b]
\centering
\caption{Registration Results on different outlier rates}\label{tab: outlier}
\begin{tabular}{lccccc}
\toprule
\multicolumn{1}{c}{Noise Type} & Outlier Ratio & \begin{tabular}[c]{@{}c@{}}MAE \\ (R/deg)\end{tabular} & \begin{tabular}[c]{@{}c@{}}MAE \\ (t/mm)\end{tabular} & \begin{tabular}[c]{@{}c@{}}CD \\ (mm)\end{tabular} &\begin{tabular}[c]{@{}c@{}}Time \\ (s)\end{tabular} \\ 
\midrule
\multirow{5}{*}{Isotropic}  & 10\%  & 0.574 & 0.456 & 0.786 &0.109\\
& 30\%    & 0.682   & 0.456  & 0.806 &0.069 \\
& 50\%    & 0.615   & 0.487  & 0.803   &0.281\\
& 70\%    & 0.682   & 0.481   & 0.853   &0.288\\
& 90\%    & \textbf{0.961}   & \textbf{0.608}  & \textbf{0.903}   &\textbf{2.320}\\ 
\midrule
\multirow{5}{*}{Anisotropic} & 10\% & 0.732 & 0.475 & 0.835 &0.081\\
& 30\%    & 0.724    & 0.471  & 0.835  &0.106\\
& 50\%    & 0.796   & 0.489  & 0.846  &0.134\\
& 70\%    & 0.899   & 0.478  & \textbf{0.875}   &0.482\\
& 90\%    & \textbf{1.104}   & \textbf{0.691}   & 0.852  &\textbf{2.714}\\ 
\bottomrule
\end{tabular}
\end{table}


\subsection{Real phantom experiments}

\begin{figure}[t]
	\centering
	\includegraphics[width=1\linewidth]{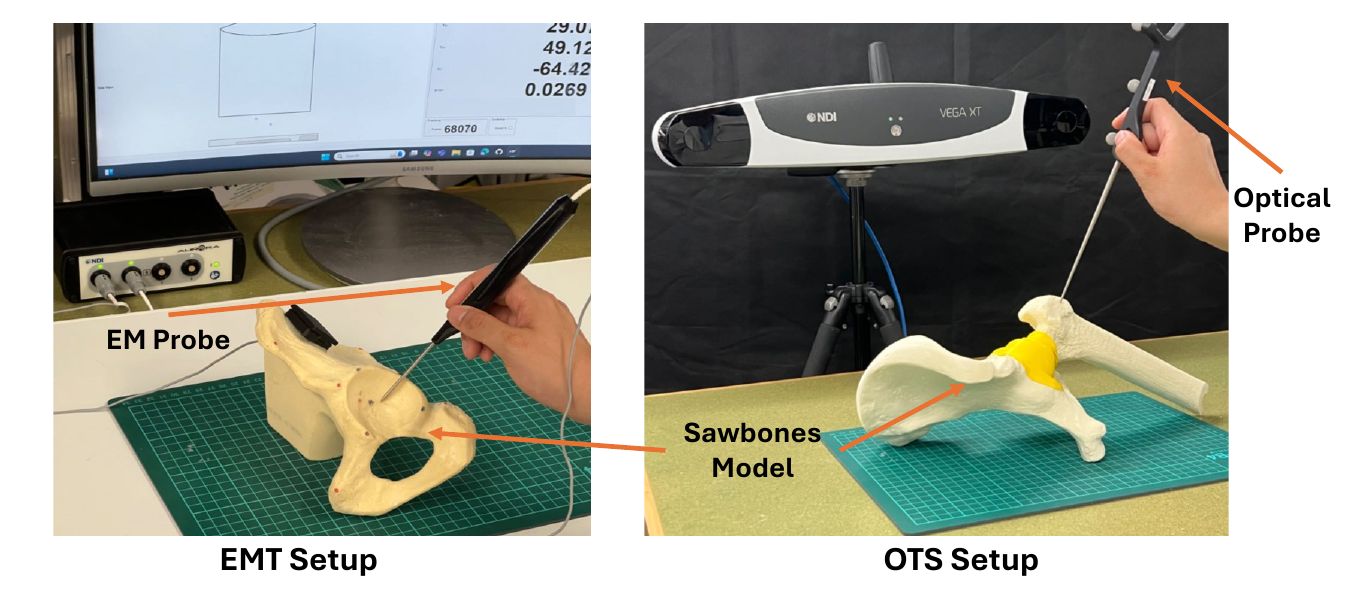}
	\caption{Setups: repeated trials were performed with the sawbones models in each of the two tracking systems. Points were collected by moving the probe over the pelvic acetabulum and proximal femur. }
	\label{fig:setup}
\end{figure}

The proposed method was also tested in phantom experiments (Fig. \ref{fig:setup}). Both optical tracking system (OTS) and electromagnetic tracking (EMT) setups were used. The OTS we used is the Polaris Vega XT (Northern Digital Inc., Canada) and the EMT is the Aurora® electromagnetic tracking (Northern Digital Inc., Canada) with tabletop field generator. The phantom experiments were performed using four hip and three femur sawbones models under each of the two tracking systems. The sawbones models were scanned in advanced by a structured light 3D scanner Solutionix C500, and the respective gradient-SDFs were generated.

\begin{figure}[h]
	\centering
	\includegraphics[width=0.9\linewidth]{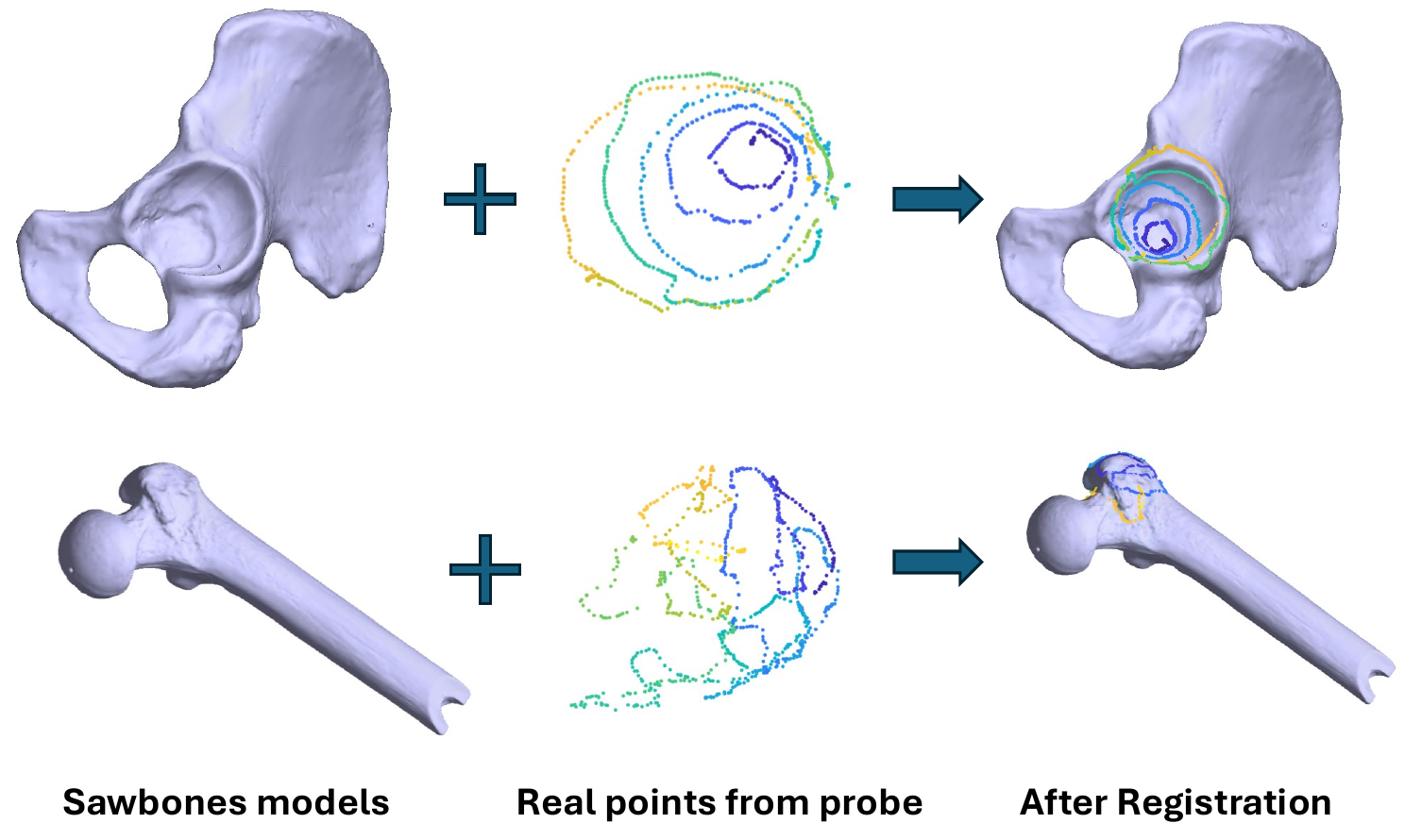}
	\caption{Two examples of the real data from phantom experiments.}
	\label{fig:real}
\end{figure}

During the experiments, the acquisition frequencies of OTS and EMT were 80 Hz and 40 Hz, respectively. The number of points taken in the pelvic acetabulum and proximal femur by the surgical probe were 600 and 1000, respectively (Fig. \ref{fig:real}). Since there is no ground truth, we use target registration error (TRE) to quantify the registration accuracy of our method. We selected 10 anatomical landmarks as targets for the \textbf{entire bone model}, in stead of the small selection in pelvic acetabulum or proximal femur. Suppose anatomical landmarks on the pre-model are defined as $P_{tar}^{pre}$. In the experiments, these landmarks are also simultaneously need to be recognised on the sawbones model and acquired by the probe, and are denoted as $P_{tar}^{intra}$. Then the TRE can be calculated as:
\begin{equation}\label{Eq:TRE}
    TRE=  \|{\hat{R} \cdot P_{tar}^{intra} + \hat{t} - P_{tar}^{pre}\|,}
\end{equation}

\noindent where $\hat{R}$ and $\hat{t}$ are the registration result. It is worth noting that these anatomical landmarks were not involved in the registration calculation. As shown in Table \ref{tab: real}, the proposed framework can achieve TRE value of 2.198 $mm$ and 2.719 $mm$ in optical and electromagnetic tracking systems, respectively, which is much more accurate than the other methods. 

\begin{table}
\centering
\caption{Registration Results in Phantom experiments}\label{tab: real}
\begin{tabular}{llcccc}
\toprule
\multirow{2}{*}{\begin{tabular}[c]{@{}l@{}} Tracking System\end{tabular}} & \multicolumn{1}{c}{\multirow{2}{*}{Methods}} & \multicolumn{2}{c}{Hip} & \multicolumn{2}{c}{Femur}  \\ \cline{3-6} & \multicolumn{1}{c}{}  & \begin{tabular}[c]{@{}c@{}}TRE \\ (mm)\end{tabular} & \begin{tabular}[c]{@{}c@{}}CD \\ (mm)\end{tabular} & \begin{tabular}[c]{@{}c@{}}TRE \\ (mm)\end{tabular} & \begin{tabular}[c]{@{}c@{}}CD \\ (mm)\end{tabular} \\ 
\midrule
\multirow{6}{*}{Optical}                                             & ICP  &  102.970  & 3.515  & 182.340  & 3.641 \\
& CPD  & 92.212 & 4.705  & 30.426 &3.962 \\
& JRMPC   &51.342 &3.327 & 35.540  &4.013   \\
& Go-ICP  &  6.782  & 0.665    & 4.563   & 0.721   \\
& Proposed   &\textbf{2.198}     &\textbf{0.505}    &\textbf{2.316}   &\textbf{0.636}    \\ 
\midrule
\multirow{6}{*}{Electromagnetic}                                              & ICP   &109.851  & 4.264 & 120.457 & 4.396   \\
& CPD     & 87.237  &  4.297  &34.225   &3.846    \\
& JRMPC     &57.325  &3.452    &37.406   &3.261    \\
& Go-ICP    & 7.314   & 0.691    & 4.713  & \textbf{1.103}  \\
& Proposed &\textbf{2.719}  &\textbf{0.662}   &\textbf{3.174}   &1.240\\ 
\bottomrule
\end{tabular}
\end{table}

\section{Discussion}
In computer-assisted orthopedic surgery, the registration of partial-to-full bone point sets presents several challenges, including partially overlapped and noise interference. We propose a novel and effective registration framework based on gradient-SDF to address these challenges. 

\textbf{Why gradient-SDF.} Gradient-SDF can pre-calculate and store the closest distance from each spatial location to the surface of geometry, and the intra-operative acquisition points in COAS only occur on the surface of bone, so it is appropriate to represent the bone model as gradient-SDF for registration. Moreover, gradient-SDF naturally contains the geometric properties that are not present in the gradients calculated by standard SDF \cite{sommer2022gradient}. So the convergence is better and faster for sparse point set without the need for correspondences.

\textbf{Initial guess.} Although the proposed framework is a correspondence-free registration method, it still requires to provide an initial guess. During our experiments, we found that our method is quite robust to inaccurate initial translation, but does require a reasonable initial rotation. In practice, a good initial rotation can be obtained by asking the surgeon to point to 3 anatomical landmarks. Compared with more than 30 points required by MAKO robot, getting a better initial value with 3 points does not add much time cost and can be easily accepted by the surgeons. 

\textbf{Limitation.} From the clinical application point of view, it would be desirable to include metrics in the current framework that allow real-time assessment of whether the surgeon has acquired enough intra-operative points. We also plan to integrate our proposed framework into the surgical robotic system \cite{walker2022proof,li2024robotic} and perform cadaver experiments in the future.

\section{Conclusion} 
This paper presents a novel and effective registration framework based on gradient-SDF to address the challenges of partial-to-full bone registration in computer-assisted orthopedic surgery. Experiments on multi-category bone structures demonstrate that our proposed framework can provide more accurate results than the state-of-the-art registration methods and be robust to 90\% outliers. The gradient-SDF results in the accurate and fast convergence for sparse and partially overlapped point set without the need for correspondences. Our method can achieve convergence in less than 1 second in real scenarios and mean target registration error values as low as 2.198 $mm$ for the entire bone model. The proposed framework enables registration to be accomplished by merely guiding the probe over the bone surface to acquire random points, thus simplifies the surgeon's intra-operative workflow and shows great potential clinical value.

\ifCLASSOPTIONcaptionsoff
  \newpage
\fi




\clearpage
\bibliographystyle{IEEEtran}
\bibliography{IEEEabrv,main}

\vfill


\end{document}